\documentclass[acmtog,screen,nonacm]{acmart}

\AtBeginDocument{%
  }

\usepackage{my}

\makeatletter
\@namedef{ver@everyshi.sty}{}
\makeatother

\begin{document}

\title{\mvd: Efficient Multiview 3D Reconstruction for Multiview Diffusion}

\author{Xin-Yang Zheng}
\authornote{Work done during internship at Microsoft.}
\email{zxy20@mails.tsinghua.edu.cn}   
\affiliation{%
  \institution{Tsinghua University}
  \city{Beijing}
  \country{P.~R.~China}
}
\author{Hao Pan}
\email{haopan@microsoft.com}   
\affiliation{%
  \institution{Microsoft Research Asia}
  \city{Beijing}
  \country{P.~R.~China}
}
\author{Yu-Xiao Guo}
\email{yuxgu@microsoft.com}  
\affiliation{%
  \institution{Microsoft Research Asia}
  \city{Beijing}
  \country{P.~R.~China}
}
\author{Xin Tong}
\email{xtong@microsoft.com}  
\affiliation{%
  \institution{Microsoft Research Asia}
  \city{Beijing}
  \country{P.~R.~China}
}
\author{Yang Liu}
\authornote{Corresponding author}
\email{yangliu@microsft.com}  
\affiliation{%
  \institution{Microsoft Research Asia}
  \city{Beijing}
  \country{P.~R.~China}
}

\authorsaddresses{}

\renewcommand{\shortauthors}{Zheng, et al.}

\begin{abstract}
  
As a promising 3D generation technique, multiview diffusion (MVD) has received a lot of attention due to its advantages in terms of generalizability, quality, and efficiency. By finetuning pretrained large image diffusion models with 3D data, the MVD methods first generate multiple views of a 3D object based on an image or text prompt and then reconstruct 3D shapes with multiview 3D reconstruction. However, the sparse views and inconsistent details in the generated images make 3D reconstruction challenging. We present {\mvd}, an efficient 3D reconstruction method for multiview diffusion (MVD) images. {\mvd} aggregates image features into a 3D feature volume by projection and convolution and then decodes volumetric features into a 3D mesh. We train {\mvd} with 3D shape collections and MVD images prompted by  rendered views of 3D shapes. To address the discrepancy between the generated multiview images and ground-truth views of the 3D shapes, we design a simple-yet-efficient \textit{view-dependent} training scheme. {\mvd} improves the 3D generation quality of MVD and is fast and robust to various MVD methods. After training, it can efficiently decode 3D meshes from multiview images within one second. We train {\mvd} with Zero-123++ and ObjectVerse-LVIS 3D dataset and demonstrate its superior performance in generating 3D models from multiview images generated by different MVD methods, using both synthetic and real images as prompts.

\end{abstract}

\begin{teaserfigure}
  \centering
\begin{overpic}[width=\textwidth]{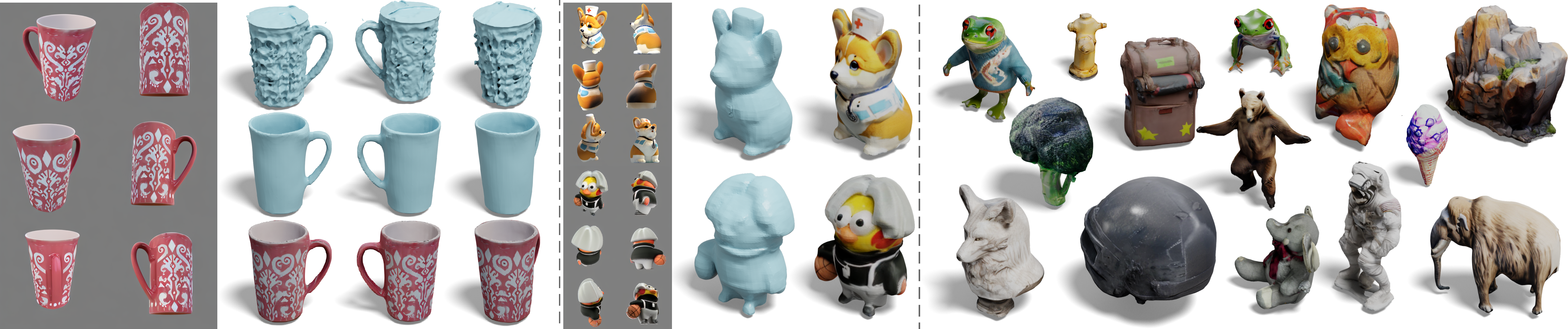}
\end{overpic}
\caption{ 
\emph{Left}: Existing 3D reconstruction methods, such as NeuS~\cite{wang2021neus}, struggle to handle the inconsistency in multiview images (in gray background) generated by multiview diffusion, leading to low-quality geometry (upper). Our {\mvd} method effectively addresses this challenge and produces more realistic geometry (middle), while being highly efficient (less than 0.5 seconds). The lower row shows our result with texture. \emph{Middle}: Visualization of two more 3D shapes reconstructed from MVD images. \emph{Right}: Diverse 3D shapes reconstructed by our method, with texture mapping.
}
\label{fig:teaser}

\end{teaserfigure}

\maketitle

\section{Introduction} \label{sec:intro}

Recently multiview diffusion (MVD) methods ~\cite{liu2023zero,shi2023mvdream,liu2023zero,Tang2023mvdiffusion} have received lots of attention as an emerging 3D generation technique. By finetuning a pre-trained image diffusion model with 3D data, the MVD methods first generate multiview images of a 3D object from text or image prompts and then produce a realistic 3D model from the generated images via multiview 3D reconstruction.
Based on the pretrained large image diffusion models, MVD methods have shown remarkable advantages over other 3D generation techniques in terms of generalizability, quality, and efficiency. However, the multiview 3D reconstruction methods used in MVD, such as NeRF~\cite{nerf} or NeuS~\cite{wang2021neus}, are designed for dense and consistent multiview input, which cannot efficiently address the sparse multiview images with inconsistent details generated by MVD methods. As a result, the reconstructed 3D shapes are often blurry and distorted (see \cref{fig:teaser}). Furthermore, the optimization process of 3D reconstruction is slow.

Several methods~\cite{liu2023syncdreamer,liu2023text,long2023wonder3d,shi2023zero123++} have been proposed to enhance the multiview consistency of MVD methods. However, these methods do not ensure pixelwise consistency across different views.
Some other methods aim to improve the 3D reconstruction quality from MVD images.
Zero-1-to-3~\cite{liu2023zero} exploits 2D diffusion priors~\cite{wang2022score} to refine both the 3D geometry and texture through costly optimization. The concurrent One-2-3-45++~\cite{liu2023oneplus} method employs a two-stage multiview-conditioned 3D diffusion model for 3D shape reconstruction. Although this method achieves better 3D reconstruction results, the discrepancy between the rendered images of 3D data used for training and the images generated by diffusion used for inference limits its performance and the 3D diffusion process remains slow.

In this paper,  we propose {\mvd}, a multiview 3D reconstruction method that reconstructs 3D shapes from multiview images generated by MVD. By noticing that the multiview images usually contain sufficient information to recover the 3D shape, we design a lightweight neural network that directly maps multiview image features to a 3D feature volume via view projection and 3D convolution, and then outputs a differentiable 3D mesh. We carefully design our network to make it robust to different view configurations of the input images. Once trained, the {\mvd} can directly decode 3D shapes from multiview images produced by various MVD methods without optimization.

Training the {\mvd} network is challenging because we do not have access to the true 3D shape that corresponds to each set of inconsistent multiview images. If we use all input views as self-supervision for training, our model will suffer from view-view inconsistency like traditional multiview 3D reconstruction methods. Alternatively, we could train our network with a collection of 3D shapes, following the MVD training procedure. For each 3D shape, we render a set of views and select one view as the prompt of MVD to generate multiview images. Then, we use either the other views or the ground-truth 3D shape as the supervision. However, this training scheme still leads to suboptimal results (shown in \cref{subsec:ablation}) due to the domain gap between the generated images (the underlying 3D shape) and the ground truth images (the training 3D shape).

By examining the discrepancy between the rendered and generated views, we observe that the discrepancy varies with the views of the generated images. Specifically, the generated view that is closer to the reference view of the prompt image is more consistent with the corresponding ground-truth image. Based on this observation, we propose a \emph{view-dependent training scheme} that enforces the inferred shape to align with the ground-truth geometry at the prompt view, and to maintain local structural similarity at other views.

We train {\mvd} with MVD images generated by Zero123++~\cite{shi2023zero123++} and the Objverse-LVIS dataset~\cite{deitke2023objaverse}, and extensively evaluate our method's performance on the unseen multiview images generated by Zero123++ and other MVD methods prompted by rendered images and real images, as well as text. We compare our method with other 3D generation methods and validate the effectiveness of our view-dependent training scheme. The experimental results show that our method significantly enhances the quality and efficiency of 3D shape reconstruction for MVD  and exhibits good generalizability for different MVD methods. \cref{fig:teaser} gathers a few results for demonstration.

In summary, we make the following contributions:
\begin{enumerate}[leftmargin=*]\setlength\itemsep{1mm}
    \item[-] We specifically address the problem of 3D reconstruction from multiview diffusion images and significantly improve the quality and efficiency of multiview diffusion for 3D generation.
    \item[-] We identify the challenges of sparsity and inconsistency with MVD images, and propose an efficient lightweight neural network trained with a view-dependent training scheme to resolve these challenges.
    \item[-] Through extensive evaluations, we show that the reconstruction model works robustly across different MVD models and complements a large family of MVD works.
\end{enumerate}
We will release our code and model to facilitate future research.
\section{Related Work}
\label{sec:related_work}

\paragraph{Optimization-based 3D generation} Numeric recent works~\cite{wang2023prolificdreamer,deng2023nerdi,qian2023magic123,chen2023fantasia3d,tang2023make,chen2023control3d,ouyang2023chasing,zeng2023ipdreamer,sun2023dreamcraft3d,melas2023realfusion,purushwalkam2023conrad}, starting from DreamField~\cite{jain2021dreamfields}, DreamFusion~\cite{poole2022dreamfusion} and score Jacobian chaining~\cite{wang2022score}, propose to generate 3D shapes from text descriptions by optimizing a 3D parametric model, such as NeRF~\cite{nerf} or deep marching tetrahedra~\cite{shen2021deep}, using pretrained CLIP model and/or 2D text-to-image models. Most of these models leverage score distillation sampling or its variants to bridge the gap between 2D and 3D domains, without requiring any 3D data for training. However, the expensive optimization hinders fast 3D generation.

\paragraph{Inference-based 3D generation} A common approach to generate 3D contents from prompts is to employ generative models such as denoise diffusion~\cite{gupta20233dgen,cheng2023sdfusion,zheng2023lasdiffusion,zhang20233dshape2vecset,li2023diffusion,liu2022iss,liu2023one,shi2023mvdream}, GANs~\cite{chen2019learning,zheng2022sdfstylegan,gao2022get3d}, and autoregressive models~\cite{autosdf2022,nash2020polygen,ibing2021octree,zhang20223dilg}. Alternatively, some regress-based methods~\cite{anonymous2023lrm,zou2023triplane,szymanowicz2023splatter,huang2023zeroshape,wu2023multiview}  map prompts to 3D outputs without probabilistic modeling, using transformer models trained on 3D or 2D data. The works mentioned above utilize different 3D output formats, such as signed distance functions (SDF), polygonal meshes, multiview images, and 3D Gaussian splatting~\cite{gaussiansplatting}.

\paragraph{Multiview diffusion} Large-scale 2D diffusion models can produce high-quality novel view synthesis by incorporating camera control mechanisms. Zero-1-to-3~\cite{liu2023zero} conditions a pretrained stable diffusion model on both the view image and the relative viewpoint change, and finetunes it on render images of Objverse~\cite{deitke2023objaverse}, a large 3D dataset. However, it requires multiple forward passes for different views. Some recent works generate multiview images with fixed viewpoints and enhance their consistency with various techniques, such as correlating intermediate images~\cite{liu2023syncdreamer,liu2023text}, aligning geometry priors~\cite{long2023wonder3d,li2023sweetdreamer,ye2023consistent,anonymous2023hifi,lu2023direct2}, strengthening cross-view attention~\cite{weng2023consistent123,yang2023consistnet}, and refining the noise schedule and local condition~\cite{shi2023zero123++,woo2023harmonyview}.

\paragraph{3D reconstruction for multiview images}
Structure-from-motion (SFM) and multiview stereo (MVS) are techniques to recover 3D geometry from multiple images (see~\cite{seitz2006comparison,ozyesil2017,Silveira2022} for surveys). They benefit from dense views and rich textures, and can use deep learning to incorporate shape priors for better robustness and quality~\cite{YAN2021106}.
Given few views and known camera poses, some MVD works~\cite{liu2023syncdreamer,long2023wonder3d} use volume rendering methods like NeuS~\cite{wang2021neus} for 3D reconstruction. One-2-3-45~\cite{liu2023one} employs SparseNeuS~\cite{long2022sparseneus} to reconstruct 3D geometry from its 2-stage multiview predictions in one pass; One-2-3-45++ trains a 3D diffusion model to convert MVD images to a signed distance function. Both methods train with ground-truth render images and ignore the inconsistency issue of MVD images.

\section{Method}
\label{sec:method}

We first formulate the problem of 3D reconstruction from MVD images (\cref{subsec:formulation}).
Then we make observations that reveal unique challenges with this problem (\cref{subsec:dilemma}).
To address the challenges, we present our training strategy and a lightweight neural network that is designed for MVD images and produces better reconstruction quality with efficiency (\cref{subsec:loss_design,subsec:decoder}).

\subsection{3D reconstruction from MVD images}
\label{subsec:formulation}

An MVD model $\mathcal{M}$ approaches 3D generation by taking an input reference image $\bI_0$ at viewpoint $v_0$ (or a text prompt) as condition, and generating images $[\bI_{i}] = \mathcal{M}(\bI_0)$ of a target 3D object at novel views $\mathcal{C}=[v_i]$.
The model is generally adapted from a pretrained large-scale image generation model (\eg stable diffusion~\cite{Rombach2022ldm}), and finetuned on multiview renderings of 3D objects from large-scale datasets~\cite{deitke2023objaverse,Objaversexl} to enhance the consistency of generated images.
Given these MVD images, a reconstruction algorithm (\eg NeuS and variants \cite{wang2021neus,long2022sparseneus}) is typically applied to obtain the final 3D object.
However, due to sparseness and lack of precise consistency of MVD images (\cref{subsec:dilemma}), typical multiview 3D reconstruction algorithms do not work  well~\cite{long2023wonder3d}.
To achieve quality and efficiency, we propose to learn a reconstruction model {\mvd} that takes $[\bI_{i}]$ as input, and recovers $\mS = \textrm{{\mvd}}([\bI_{i}])$ as output.
The problem remains of how to supervise $\mS$, which reveals a unique challenge for MVD reconstruction, as discussed next.

\subsection{The MVD reconstruction dilemma}
\label{subsec:dilemma}

\paragraph{Observations about MVD images}
Despite the differences of MVD models in technical details (Sec.~\ref{sec:related_work}), we make the following observations common to their generated images:
\begin{enumerate}[leftmargin=*]\setlength\itemsep{1mm}
    \item[\textbf{P1}] Viewpoints of MVD images are known and sparsely scattered.
    \item[\textbf{P2}] MVD images tend to be consistent in 3D but lack precision.
    \item[\textbf{P3}] The closer to the input view, the better generated images match the training object.
\end{enumerate}

\textbf{P1} derives from the formulation of MVD.
In particular, viewpoints of generated images are specified explicitly, which saves the trouble of pose estimation for 3D reconstruction algorithms.
The sparsity of MVD images is a result of limited computational power.
MVD models are trained to generate a limited number (\eg 4-16) of images simultaneously with enhanced consistency; beyond that, the memory and computational cost become unbearable~\cite{shi2023mvdream,long2023wonder3d}.

\textbf{P2} also follows naturally from MVD approaches. Indeed, while MVD models strive to enhance multiview consistency by diverse techniques of cross-view modulation \cite{shi2023zero123++,liu2023syncdreamer,shi2023mvdream}, there is no guarantee of pixelwise 3D consistency of the generated images.

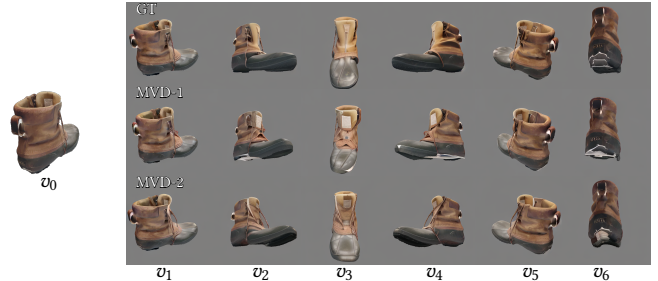
\begin{figure}
    \centering
    \begin{overpic}[width=1\linewidth]{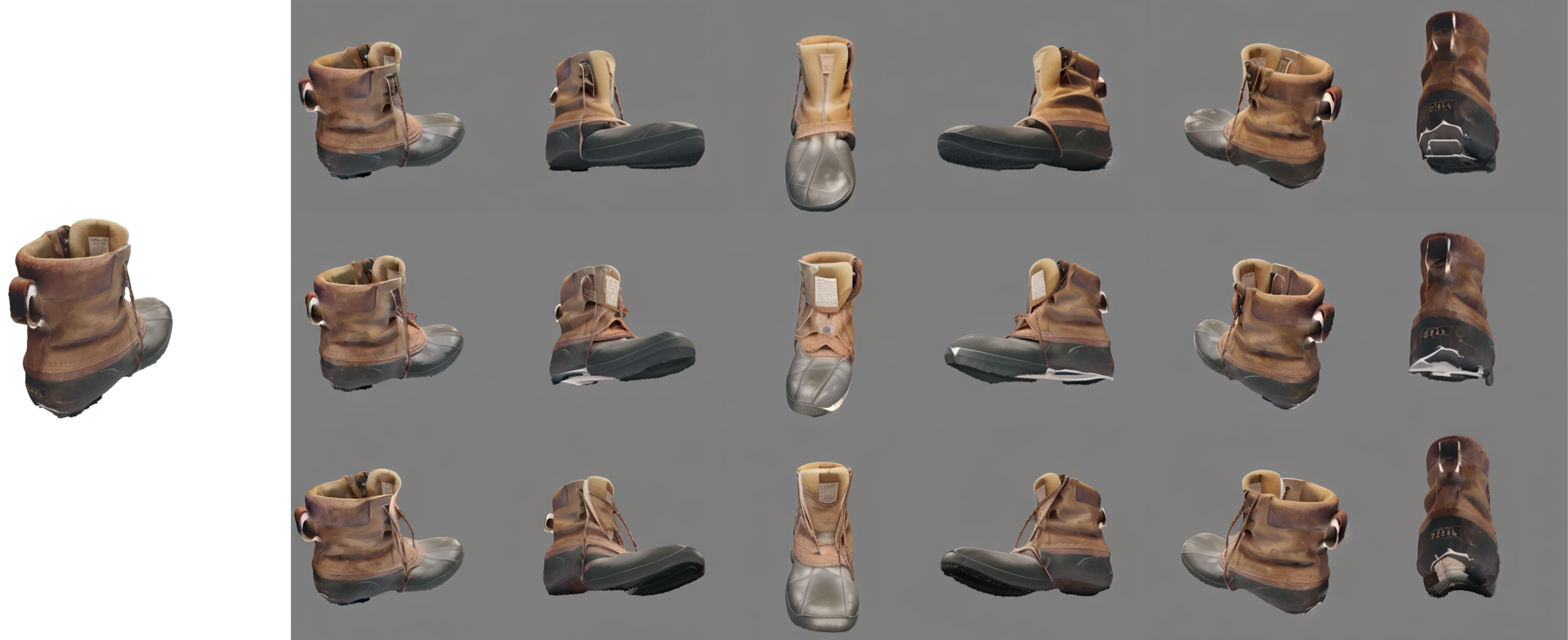}
    \put(23,-2){\small $v_1$}
    \put(38,-2){\small $v_2$}
    \put(51,-2){\small $v_3$}
    \put(65,-2){\small $v_4$}
    \put(80,-2){\small $v_5$}
    \put(91,-2){\small $v_6$}
    \put(5,12){\small $v_0$}
    \put(20,39){\scriptsize \contour{black}{\textcolor{white}{GT}}}
    \put(20,26){\scriptsize \contour{black}{\textcolor{white}{MVD-1}}}
    \put(20,12){\scriptsize \contour{black}{\textcolor{white}{MVD-2}}}
    \end{overpic}
    \caption{Inconsistency from training object increases as the viewpoint moves away from the reference image.
    }
    \label{fig:inconsistency} \vspace{-3mm}
\end{figure}

\textbf{P3} presents an observation that can be confirmed by examining \cref{fig:inconsistency}. In this figure, $v_0$ represents a rendered view of a 3D object used for training, which serves as the input to an MVD model --- Zero-123++~\cite{shi2023zero123++}. This model generates images from various viewpoints, denoted as $v_1$ to $v_6$. It is worth noting that $v_1$ and $v_6$ are closer to $v_0$. We generated two sets of MVD images: MVD-1 and MVD-2. Among these sets, the images at $v_1$ and $v_6$ exhibit a higher similarity to the ground-truth images (rendered in the first row) compared to the images at other viewpoints. We also quantitatively validated the observation on 240 3D models (see \cref{tab:stat}), by comparing the image differences between the generated and GT views using PSNR metric. 

Taken together, these observations present the following unique challenge for 3D reconstruction from MVD images.

\begin{table}[t]
    \caption{Quantitative evaluation of image differences between generated views and GT views, averaged on 240 objects. 
    } \label{tab:stat}
    \centering
    \vspace{-3mm}
    \begin{tabular}{@{}l*{6}{c}@{}}
        \toprule
                               & $v_1$          & $v_2$ & $v_3$ & $v_4$ & $v_5$ & $v_6$          \\ \midrule
        \thead{PSNR}$\uparrow$ & \textbf{25.61} & 23.57 & 23.37 & 23.39 & 23.77 & \textbf{25.01} \\
        \bottomrule
    \end{tabular}

\end{table}

\paragraph{The MVD reconstruction dilemma}
We note there is a dilemma with 3D reconstruction from MVD images.
On one hand, there is no ground-truth 3D shape to supervise the training of the reconstruction model, because as noted in \textbf{P3}, the MVD images at viewpoints away from the reference view deviate from the training object $\overline{\mS}$, and to make $\mS$ approach $\overline{\mS}$ would contradict with the input images $[\bI_{i}]$.
On the other hand, directly comparing the rendered images with $[\bI_{i}]$ would also be problematic, since according to \textbf{P2}, $[\bI_{i}]$ are not precisely consistent in 3D.

We validate the dilemma through experiments in Sec.~\ref{subsec:ablation}, where the contradictory situations are shown to produce suboptimal results.
To solve this dilemma of input-output mismatch, we use 3D shapes as proxies, construct training inputs through image-prompted MVD models, and design loss functions that avoid the dilemma.
Note that while the training relies on image-prompted MVD models, at test time our model naturally extends to processing MVD images of text-prompted models.

\subsection{View-dependent training}
\label{subsec:loss_design}

For each object $\overline{\mS}$, we render it in a random viewpoint as the reference image $\bI_0$, and in the specified viewpoints $[v_i]$ the proxy images $[\overline{\bI_i}]$.
Correspondingly, $[\bI_i] = \mathcal{M}(\bI_0)$ is the set of MVD images to be used as input for {\mvd}.
We formulate the reconstruction loss as follows:
\begin{equation} \label{eq:obj}
  \min \sum_{i=0}^N \dis\left([\pi_i(\mS), \overline{\bI_i}], v_i\right),
\end{equation}
where $\pi_i(\mS)$ is the differentiable rendering of $\mS$ at the viewpoint $v_i$, and $\dis([\cdot,\cdot],v_i)$ is a view-dependent loss function that quantifies the discrepancy between two images differently according to the viewpoint $v_i$.
Specifically, according to \textbf{P3}, at $v_0$ the reference view, we expect the recovered shape can fully match $\overline{\bI_0}$ in pixelwise details, while at the other viewpoints, we only ask for structural similarity with $\overline{\bI_i}, i\neq 0$.
Therefore, we have
\begin{equation}
  \dis([\mathbf{x},\mathbf{y}], v_i) =
  \begin{cases}
    \mathcal{L}_{pixel} (\mathbf{x}, \mathbf{y}) & \text{for view }v_0          \\
    \mathcal{L}_{LPIPS} (\mathbf{x}, \mathbf{y}) & \text{for view }v_i, i\neq 0
  \end{cases}
\end{equation}
Here, ${L}_{LPIPS}$ loss measures perceptual patch similarity between two images using pretrained image backbones~\cite{zhang2018unreasonable}.

\begin{figure*}[t]
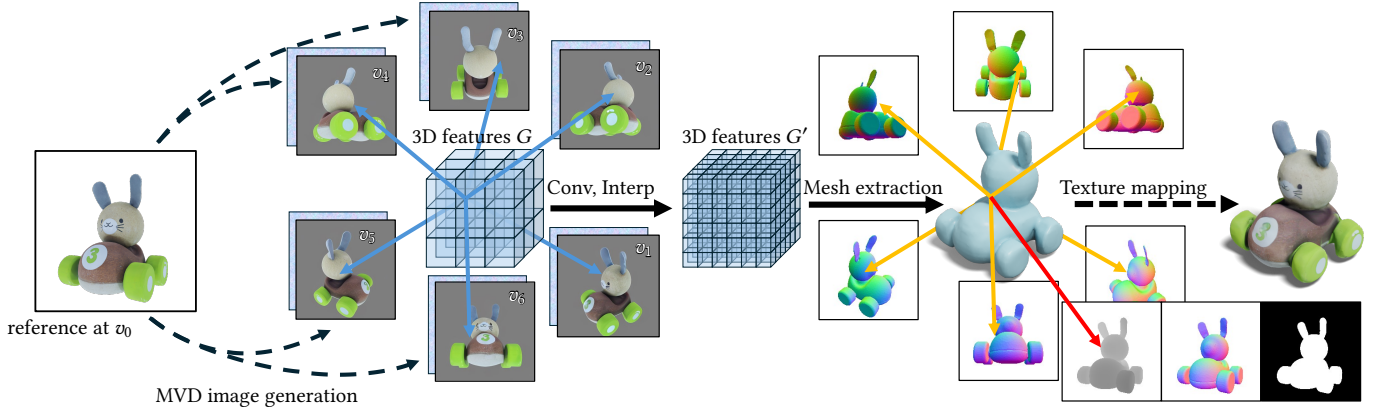

    \centering
    \begin{overpic}[width=\linewidth]{images/mvd_pipeline_dnm_maps}
    \put(-2, 5){\small reference at $v_0$}
    \put(44.5, 11){\small \contour{black}{\textcolor{white}{$v_1$}}}
    \put(35.1, 7.5){\small \contour{black}{\textcolor{white}{$v_6$}}}
    \put(24, 12){\small \contour{black}{\textcolor{white}{$v_5$}}}
    \put(25, 24){\small \contour{black}{\textcolor{white}{$v_4$}}}
    \put(35, 27){\small \contour{black}{\textcolor{white}{$v_3$}}}
    \put(44.5, 24.5){\small \contour{black}{\textcolor{white}{$v_2$}}}
    \put(9, 0){\small MVD image generation}
    \put(28, 19.2){\small 3D features $G$}
    \put(38, 15.4){\small Conv, Interp}
    \put(48, 19.2){\small 3D features $G'$}
    \put(57, 15.4){\small Mesh extraction}
    \put(76, 15.4){\small Texture mapping}
    \end{overpic}
    \caption{
    Method overview.
       The MVD model produces a set of images from different viewpoints based on a reference image. {\mvd} extracts and averages features from these images for each point in a coarse 3D grid $G$, and interpolates them into a finer grid $G'$, from which the surface mesh is extracted in a differentiable manner. The mesh reconstruction during  training is supervised with pixelwise loss (red arrow) against depth/normal/mask maps at the reference view $v_0$, and with structural loss (yellow arrow) against normal maps at the other views. The reconstructed mesh can be textured by mapping to MVD images.
    }
    \label{fig:overview}
\end{figure*}

Disentangled reconstruction of geometry and texture generally leads to better qualities in both geometry and texture \cite{chen2023fantasia3d}.
In this work, we follow this approach and focus on recovering the shape geometry, and apply multiview texture mapping on the reconstructed detailed shapes subsequently (\cref{subsec:training}).
To supervise the geometry reconstruction, we rewrite the $v_0$ pixelwise losses as:
\begin{align}
  \mathcal{L}_{\mathrm{d}}\left(\mathbf{d}, \overline{\mathbf{d}}\right) & = \sum_{p}{\frac{\left\| \mathbf{d}(p) - \overline{\mathbf{d}}(p) \right\|_1}{\overline{\mathbf{d}}(p) - d_{min}}}, \\
  \mathcal{L}_{\mathrm{n}}\left(\mathbf{n}, \overline{\mathbf{n}}\right) & = \sum_{p}{1 - \left|\mathbf{n}(p)\cdot \overline{\mathbf{n}}(p)\right|},                                           \\
  \mathcal{L}_{\mathrm{m}}\left(\mathbf{m}, \overline{\mathbf{m}}\right) & = \left\|\mathbf{m}(p) - \overline{\mathbf{m}}(p)\right\|_2^2
\end{align}
for depth, normal and mask maps $\mathbf{d},\mathbf{n},\mathbf{m}$ respectively.
Here $p$ iterates over the pixels, and $d_{min}$ is the minimum depth value of bounding sphere, subtracted to normalize the arbitrary camera translations of MVD models.

For views other than $v_0$, we apply the structure similarity loss on normal maps.
Therefore, the total loss is
\begin{equation}
  \begin{aligned}
    \mathcal{L} = & \quad \lambda_1 \mathcal{L}_{\mathrm{d}}(\mathbf{d}_{v_0}, \overline{\mathbf{d}}_{v_0}) + \lambda_2 \mathcal{L}_{\mathrm{n}}\left(\mathbf{n}_{v_0}, \overline{\mathbf{n}}_{v_0}\right) + \lambda_3 \mathcal{L}_{\mathrm{m}}(\mathbf{m}_{v_0}, \overline{\mathbf{m}}_{v_0}) \\&+ \lambda_4\sum_{v_i,i\neq 0}{\mathcal{L}_{\mathrm{LPIPS}}(\mathbf{n}_{v_i}, \overline{\mathbf{n}}_{v_i})}+  \mathcal{L}_{\mathrm{reg}}
  \end{aligned}
\end{equation}
where $\mathcal{L}_{\mathrm{reg}}$ is a regularization loss for 3D mesh (\cref{subsec:decoder}).
We empirically set $\lambda_1 = \lambda_3 = 1.0, \lambda_2= 0.2, \lambda_4= 0.1$.
Note that while for each training sample only the reference view provides strong pixelwise guidance and the other views regulate shape structures, we have synthesized training samples with random reference views (Sec.~\ref{subsec:training}), which ensures that the trained model can recover accurate geometry at arbitrary views.

\subsection{Reconstruction model}
\label{subsec:decoder}

To complement a wide range of MVD models that differ in aspects, including image viewpoints and resolutions, we design {\mvd} to be independent of specific MVD models and accommodate diverse sets of sparse images in a generalizable way.
In particular, the {\mvd} model recovers 3D features by fetching image features according to given viewpoints, and after minimal local transformations that are invariant to image numbers produces high-quality 3D geometry in a mesh representation.
We illustrate the overview of our approach in \cref{fig:overview}, and expand on the model details next.

\paragraph{2D-to-3D feature transformation}

To ensure the independence of specific MVD models, we use a pre-trained DINOv2 model \cite{oquab2023dinov2}  to turn the MVD images $[\bI_i]$ into 2D feature maps. The original resolution of DiNOv2 feature maps is $37\times 37$, we upsample the feature maps to $64\times 64$ resolution, and fabricate them via view-shared 2D convolution. The resulting feature maps are denoted by $[\mathbf{F}_i\in \mathbb{R}^{D_{2d}}]$.
To obtain 3D pointwise features from these 2D feature maps, given a spatial point $p \in G$ where $G$ is a regular grid of resolution $R_G^3$ tessellating the unit cubic space, we fetch image features for $p$ by $[\mathbf{f}_{p, i} = \mathbf{F}_i\left(\pi_i(p)\right)]$ through specified view projection, where $\mathbf{F}(\cdot)$ queries the image feature map through bilinear interpolation.
To remain invariant to the number and ordering of images, we use average pooling to fuse the image features into a single feature vector, \ie $[\mathbf{f}_p = \textrm{avg}\left(\mathbf{f}_{p, i}\right)]$, similar to the design of PointNeRF~\cite{yu2020pixelnerf} in supporting multiviews.
The 3D feature grid is then transformed by a small number of 3D convolution layers before extracting the mesh representation.
Overall, the network model is lightweight, comprising a few layers and parameters that we find sufficient for reconstruction from the MVD images.  It is important to note that the image view at $v_0$ (\cref{subsec:loss_design}) is solely utilized in the loss computation during training phase, and it is not required by the network for inference.

\paragraph{Shape representation}

We choose triangle meshes as our shape representation for their compactness and suitability as 3D assets, and adopt FlexiCubes~\cite{flexicubes} to produce mesh output with differentiability. At the core of FlexiCubes is to transform a learned feature grid to signed distance and deformation functions, from which the isosurface mesh is extracted by marching cubes.
In particular, we use a higher resolution grid $G'$ of shape $R_{G'}^3$ as the defining grid of Flexicubes, whose grid point features are interpolated trilinearly from the 3D feature grid $G$.
For each grid point $p\in G'$, we obtain its signed distance value and deformation parameters by learned mappings implemented as shallow MLPs.  We adopt the regularizers of Flexicubes to define our $\mathcal{L}_{\mathrm{reg}}$, which reduce unnecessary oscillations of both distance fields and surface meshes.

\section{Experimental Analysis} \label{suc:results}
We present a thorough evaluation of {\mvd} in this section. \cref{subsec:training} details the training setup and the evaluation protocol. \cref{subsec:eval} shows quantitative and qualitative results, and compares {\mvd} with other methods. \cref{subsec:generalizability} examines the applicability of {\mvd} to other MVD models. \cref{subsec:ablation} explains the rationale behind the training scheme of {\mvd}. \cref{subsec:limitation} discusses the limitations of our approach.

\subsection{Experiment setup} \label{subsec:training}

\paragraph{Training data preparation}
We adopt Zero-123++~\cite{shi2023zero123++}, a state-of-the-art MVD model that is conditioned on an input image and generate six views whose azimuths are relative to the input view while having fixed elevations, to prepare MVD images for training. We use Objaverse-LVIS~\cite{deitke2023objaverse}, a curated subset of Objaverse with diverse and textured 3D objects, as our 3D training data. For each object, we render three images from random views with a resolution of $512 \times 512$, and for each view image, we generate six groups of multiview images by Zero-123++.
The MVD images have a resolution of $320\times 320$. We also use the NVdiffrast library~\cite{laine2020modular} to render the depth and normal maps of the 3D object at the same views including views of MVD images, with the image resolution of $512 \times 512$.

\paragraph{Network setting}
The trainable parameters of {\mvd} consist of four 2D convolutional layers (channel dims: $[768,512,256,128,32]$, kernel size: $3$) and four 3D convolutional layers (channel dims: $[32,32,32,32,32]$, kernel size: $3$), each with a residual block structure, and a three-layer MLP (channel dims: $[32,64,64,4]$) after interpolating the 3D features, in total \SI{20}{M} parameters.
Correspondingly, we have $D_{2d} = 128$, $R_G = 32$ and $R_{G'} = 80$.
We trained the model for \SI{100}{k} steps on eight NVIDIA  (\SI{16}{G}) V100 GPUs, with batch size $1$ per GPU, taking around 12 hours.

\paragraph{Network inference} {\mvd} takes MVD images with known viewpoints as inputs for network inference. It achieves high efficiency: the DINOv2 feature extraction takes \SI{0.2}{seconds}, and the network's forward pass also takes \SI{0.2}{seconds} with a peak GPU-memory usage of \SI{5.2}{GB} on an NVIDIA GeForce RTX 3090 GPU. This is much faster than NeuS, which takes \SI{15}{minutes}.

\paragraph{Evaluation protocol} For evaluating geometry quality of reconstructed shapes, we use the \emph{Google Scan Objects} (GSO) dataset \cite{downs2022google} as the test set, since it is not used for training existing MVD models and {\mvd}. We select 50 diverse objects from GSO for evaluation and comparison. For each object, we render a random view as input for different image-conditioned MVD models and image-to-3D models. After 3D reconstruction by our method or other competing methods, we render the object from \SI{32} spatially uniform views, obtaining depth and normal maps. We compute PSNR, SSIM (Structural similarity~\cite{SSIM}), LPIPS (learned perceptual image patch similarity~\cite{zhang2018unreasonable}) metrics of these geometric maps against the ground-truth views rendered from the original 3D object. We also evaluate the Chamfer distance (CD) and Earth mover's distance (EMD), using \SI{2048} uniformly sampled points.  The subscripts $_d$ and $_n$ are employed to indicate that the metric is calculated on the depth map and normal map, respectively. To improve readability, the values of CD, EMD, SSIM, and LPIPS are all multiplied by 100.
However, this evaluation protocol measures the reconstruction quality only in relation to the reference object, ignoring the diversity of MVD images from the reference. Therefore, visual inspection is also essential to assess the shape quality and realism.  

\paragraph{Texture mapping} We develop a simple algorithm to convert multiview images to UV-map for texture mapping after reconstructing the geometry. The algorithm consists of four steps: (1) \emph{color initialization}: we assign a color to each surface point by selecting the view that has the largest visible area of the projected triangle and using its texture coordinates; (2) \emph{color blending}: we smooth the color transitions by averaging the colors of a small neighborhood around each surface point; (3) \emph{color filling}: we fill in the gaps where no color is assigned due to occlusion by propagating the colors from nearby regions. This algorithm is fast: takes \SI{0.5}{seconds} for one shape. However, it is only for visualization purposes as shown in \cref{fig:teaser} and does not address the inconsistency of MVD images or inpaint the occluded regions with better image content. We defer these challenges for future work.

\subsection{Performance evaluation} \label{subsec:eval}

\begin{table*}[t]
    \caption{Quantitative evaluation of single-view 3D reconstruction on GSO dataset~\cite{downs2022google}. \emph{Recon.} refers to the reconstruction method for MVD images. $N_I$ denotes the number of MVD images used for 3D reconstruction.
    } \label{tab:neus_vs_mvd}
    \centering
    \scalebox{1}{
        \begin{tabular}{@{}lcc|c*{8}{c}@{}}
            \toprule
            \thead{MVD Model}                                                       & $N_I$ & \thead{Recon.} & \thead{CD}$\big\downarrow$ & \thead{EMD}$\big\downarrow$ & \thead{PSNR}$_d$$\big\uparrow$ & \thead{SSIM}$_d$$\big\uparrow$ & \thead{LPIPS}$_d$$\big\downarrow$ & \thead{PSNR}$_n$$\big\uparrow$ & \thead{SSIM}$_n$$\big\uparrow$ & \thead{LPIPS}$_n$$\big\downarrow$
            \\ \midrule
            \rowcolor{gray!10}                                                      & 6     & NeuS           & 1.543                      & 16.02                       & 21.62                          & 87.01                          & 15.52                             & 16.39                          & 72.89                          & 22.49                             \\
            \rowcolor{gray!10} \multirow{-2}{*}{Zero-123++~\cite{shi2023zero123++}} & 6     & {\mvd}         & \textbf{1.044}             & \textbf{13.58}              & \textbf{23.31}                 & \textbf{89.20}                 & \textbf{10.80}                    & \textbf{18.28}                 & \textbf{80.01}                 & \textbf{16.34}                    \\ \midrule
                                                                                    & 6     & NeuS           & 2.146                      & 19.95                       & 19.91                          & 84.37                          & 20.31                             & 14.91                          & 69.65                          & 27.24                             \\
            \multirow{-2}{*}{Stable Zero123~\cite{stablezero123}}                   & 6     & {\mvd}         & 2.631                      & 18.66                       & 20.98                          & 86.88                          & 15.13                             & 16.34                          & 77.04                          & 21.38                             \\ \midrule
            {SyncDreamer~\cite{liu2023syncdreamer}}                                 & 16    & NeuS           & 2.082                      & 19.19                       & 20.42                          & 86.18                          & 15.44                             & 15.87                          & 75.29                          & 22.99                             \\
            {Wonder3D~\cite{long2023wonder3d}}                                      & 6     & NeuS           & 2.347                      & 21.54                       & 19.68                          & 85.45                          & 18.10                             & 15.30                          & 74.59                          & 25.50                             \\
            One-2-3-45 \cite{liu2023one}                                            & 36    & SparseNeuS     & 5.121                      & 26.82                       & 17.86                          & 83.83                          & 22.13                             & 13.90                          & 70.42                          & 30.26                             \\ \midrule
            Shap-E \cite{jun2023shap}                                               & -     & -              & 4.553                      & 24.48                       & 19.18                          & 84.48                          & 18.83                             & 14.83                          & 73.31                          & 26.05                             \\
            LRM \cite{anonymous2023lrm}                                             & -     & -              & 1.716                      & 17.12                       & 20.34                          & 86.04                          & 15.28                             & 15.61                          & 69.65                          & 24.04                             \\
            \bottomrule
        \end{tabular}
    }
\end{table*}

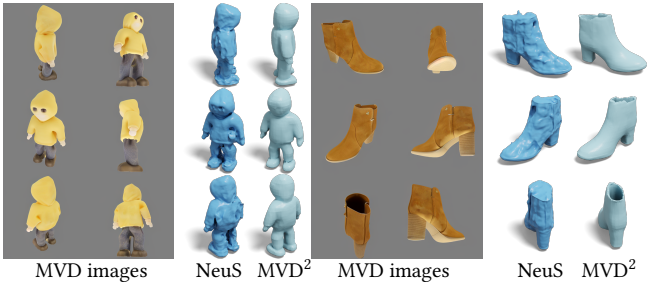
\begin{figure}[t]
    \centering
    \begin{overpic}[width=\linewidth]{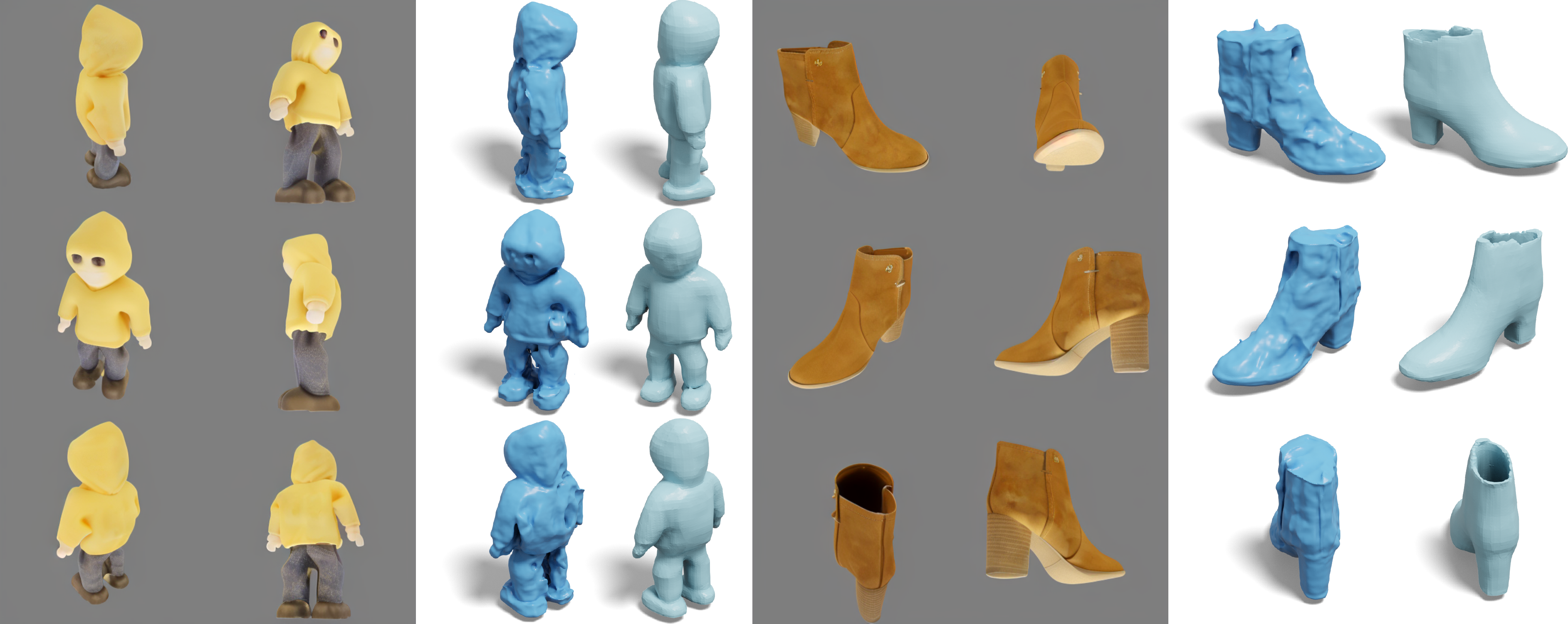}
    \put(5,-2.8){\small MVD images}
    \put(52,-2.8){\small MVD images}
    \put(30,-2.8){\small NeuS}
    \put(39.5,-2.8){\small {\mvd}}
    \put(80,-2.8){\small NeuS}
    \put(90,-2.8){\small {\mvd}}
    \end{overpic}
    \caption{3D reconstruction of Zero123++'s MVD images. The results of NeuS and {\mvd} are rendered in blue and cyan tones, respectively, from three different views.}
    \label{fig:vs_neus} \vspace{-3mm}
\end{figure}

\paragraph{Quantitative evaluation}
Our {\mvd} method outperforms NeuS in recovering 3D geometry from MVD images generated by Zero-123++, as shown in the first row of \cref{tab:neus_vs_mvd}. The higher SSIM values and lower LIPIPS values indicate that {\mvd} produces geometry that is more structurally similar to GSO data. \cref{fig:vs_neus} shows visual comparisons on two examples. NeuS produces more distorted geometry because it does not account for multiview inconsistency. On the other hand, {\mvd}'s results are more visually pleasing.

\paragraph{Robustness} The pretrained MVD can also reconstruct MVD images generated by other MVD models using the same viewpoints. We use Stable Zero123~\cite{stablezero123}, an enhanced version of Zero-1-to-3 model~\cite{liu2023zero} that supports arbitrary viewpoints for view synthesis. We employ the same camera setup as Zero-123++ and generate MVD images. We reconstruct 3D shapes from MVD images, using both NeuS and {\mvd}. The second row of \cref{tab:neus_vs_mvd} shows that {\mvd} outperforms NeuS significantly in most metrics except CD.

\paragraph{Comparison with MVD-based methods}
We also compare with other recent MVD methods: SyncDreamer~\cite{liu2023syncdreamer}, Wonder3D~\cite{long2023wonder3d}, One-2-3-45~\cite{liu2023one}. SyncDreamer generates 16 views with uniformly sampled azimuth angles and a fixed elevation angle of \SI{30}{\degree}, while Wonder3D generates 6 orthogonal side views. We use their default NeuS-based implementation for 3D reconstruction.  One-2-3-45 uses Zero-1-to-3 to generate 36 view images and a pretrained SparseNeuS for 3D reconstruction. The quantitative report presented in the third row of \cref{tab:neus_vs_mvd} indicates that they perform significantly worse than Zero-123++ with either our {\mvd} or NeuS, and Stable Zero123 with {\mvd}.
We also test a concurrent method --- One-2-3-45++~\cite{liu2023oneplus} that generates 3D shapes using MVD-image-conditioned 3D diffusion, where MVD images are generated by an MVD model identical to Zero-123++. Since the code of this method is not released, we only use its commercial web service to generate a few samples for visual evaluation.

\paragraph{Comparison with image-to-3D methods} We further compare our approach with recent image-to-3D approaches that do not use multiviews, including Shap-E~\cite{jun2023shap}, which employs a conditional diffusion model to generate a 3D implicit function, and LRM~\cite{anonymous2023lrm}, which maps the input image to a NeRF by a large transformer model. Since the official implementation of LRM is not available, we use a third-party implementation~\cite{openlrm} for comparison. The results are shown in the third row of \cref{tab:neus_vs_mvd}. Both LRM and Shap-E are inferior to Zero-123++ with {\mvd} and Stable Zero123 with {\mvd}.

\begin{figure*}[t]
    \centering
    \begin{overpic}[width=0.9\linewidth]{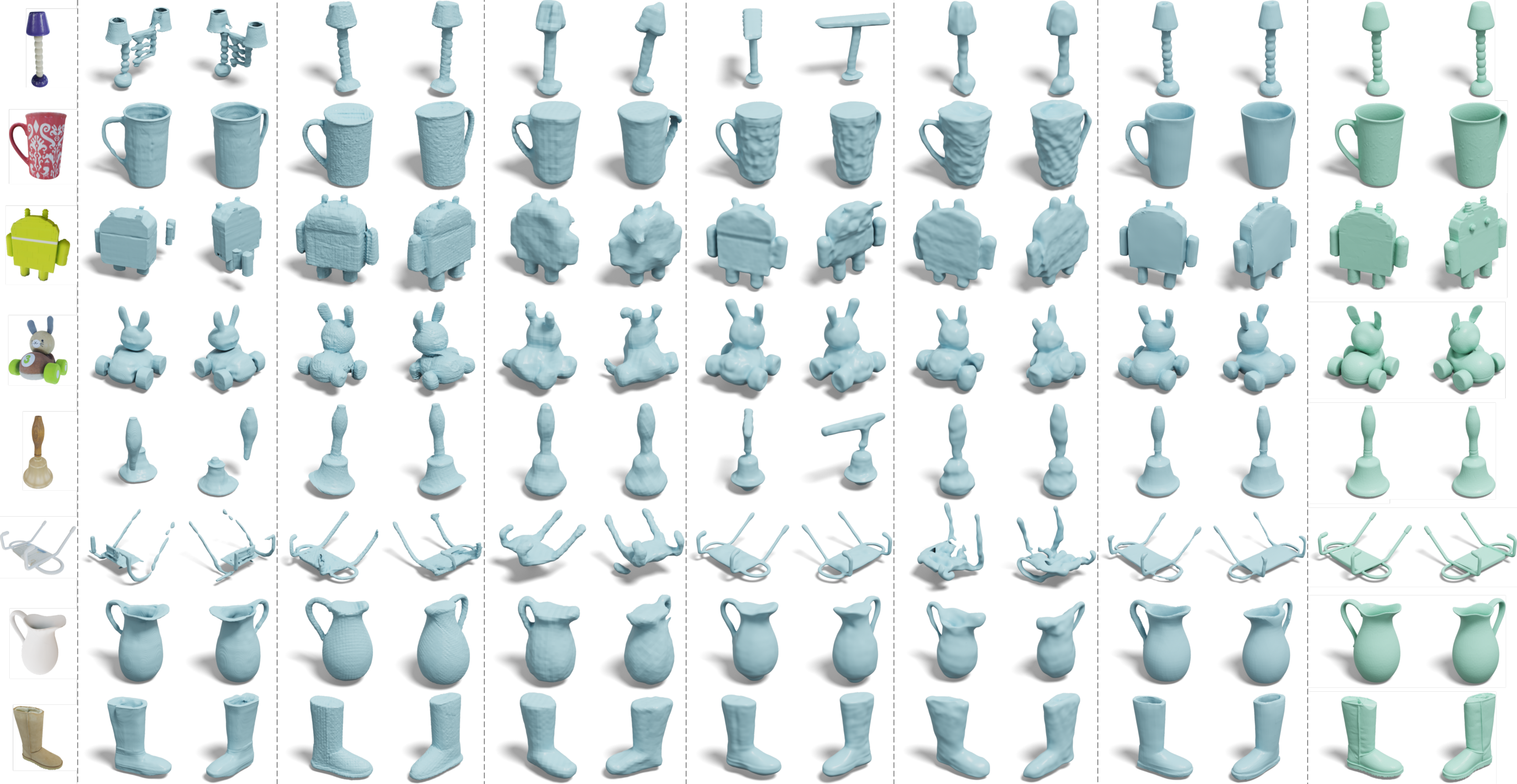}
    \put(1,  -1.5){\small Input}
    \put(10, -1.5){\small Shap-E}
    \put(24, -1.5){\small LRM}
    \put(35, -1.5){\small One-2-3-45}
    \put(48, -1.5){\small SynDreamer}
    \put(62, -1.5){\small Wonder3D}
    \put(77.5, -1.5){\small Ours}
    \put(92.1, -1.5){\small GT}
    \end{overpic}
    \caption{Visual comparison of differential approaches in single-view 3D generation on the GSO dataset. For each result, we render two different views for visualization. \textbf{Ours} refers to using {\mvd} with Zero-123++.
     The two rightmost images are rendered views of the referenced 3D object.      }
    \label{fig:visual_gso}
\end{figure*}

\begin{figure*}[h]
    \centering
    \begin{overpic}[width=0.9\linewidth]{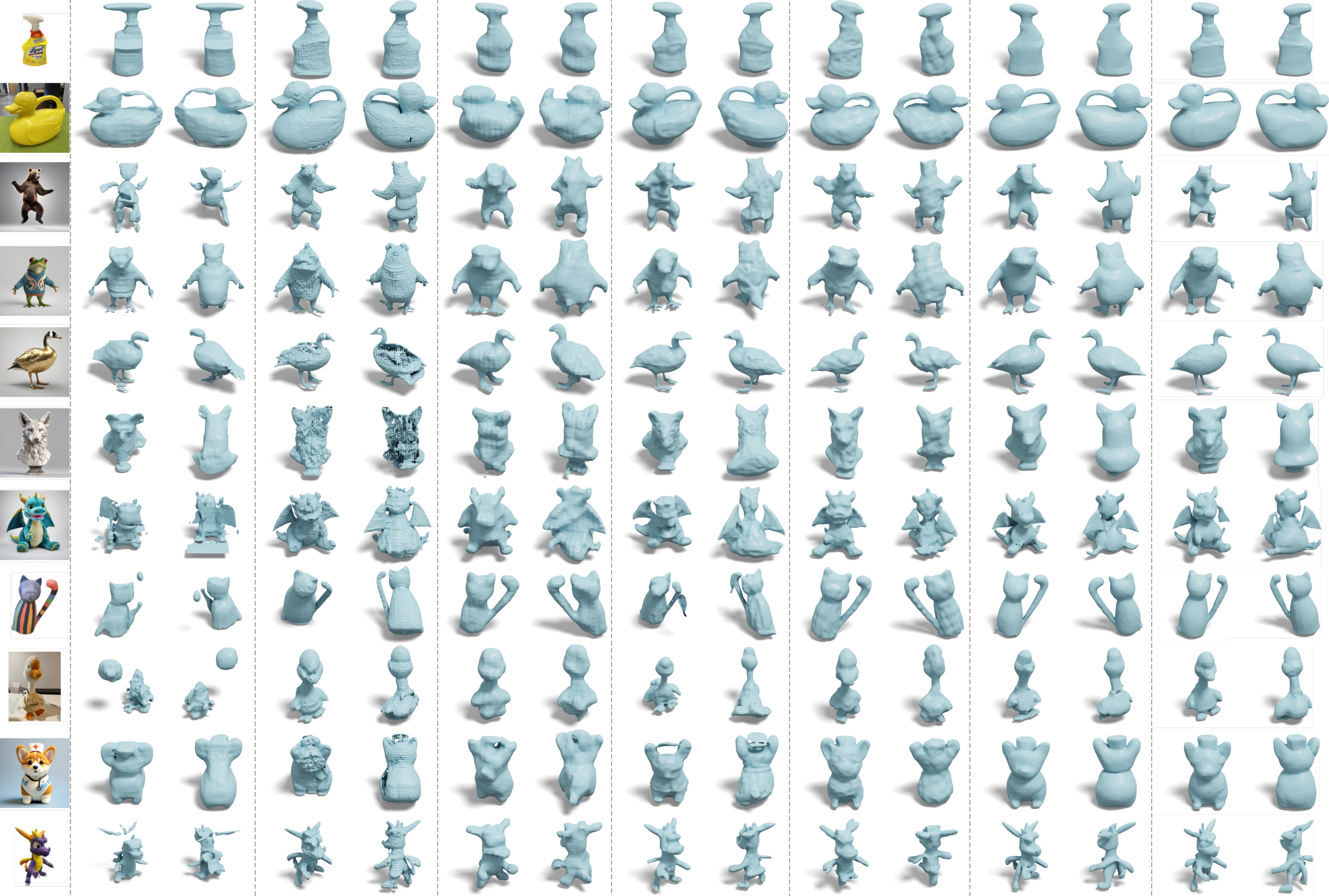}
    \put(1,-1.5){\small input}
    \put(10,-1.5){\small Shap-E}
    \put(25,-1.5){\small LRM}
    \put(36,-1.5){\small One-2-3-45}
    \put(48,-1.5){\small SyncDreamer}
    \put(62,-1.5){\small Wonder3D}
    \put(75,-1.5){\small One-2-3-45++}
    \put(92,-1.5){\small Ours}
    \end{overpic}
    \caption{visual comparison of 3D generation conditioned on Internet images. The image background is removed before feeding to different methods. }
    \label{fig:visual_novel}
\end{figure*}

\paragraph{Visual comparison} We select a subset of rendering images from the GSO dataset and random images from the Internet to use as inputs to the above compared methods. The reconstructed 3D shapes are compared visually in \cref{fig:visual_gso,fig:visual_novel}. The visualizations demonstrate that our approach produces more visually compelling results with finer geometric details.

\paragraph{Inference efficiency} On the single-image-to-3D task,  we need 12 seconds to generate six views (in 75 diffusion steps, using Zero-123++) and less than 1 second to produce the final textured mesh via {\mvd}. This is comparable to Shape-E, which takes 12 seconds to generate the 3D result, but slower than LRM, which only needs 5 seconds.  The other methods compared are much slower due to MVD generation, NeuS optimization, or 3D diffusion.

\subsection{Model generalizability} \label{subsec:generalizability}
As {\mvd} extracts image features based on view positions and applies mean pooling to combine features from different views, it can accommodate varying numbers and viewpoints, in theory. Hence, we examine whether {\mvd} pretrained on Zero-123++'s output can adapt to MVD images generated by other MVD models, with different view settings.

\begin{figure*}[t]
    \centering
    \includegraphics[width=0.9\linewidth]{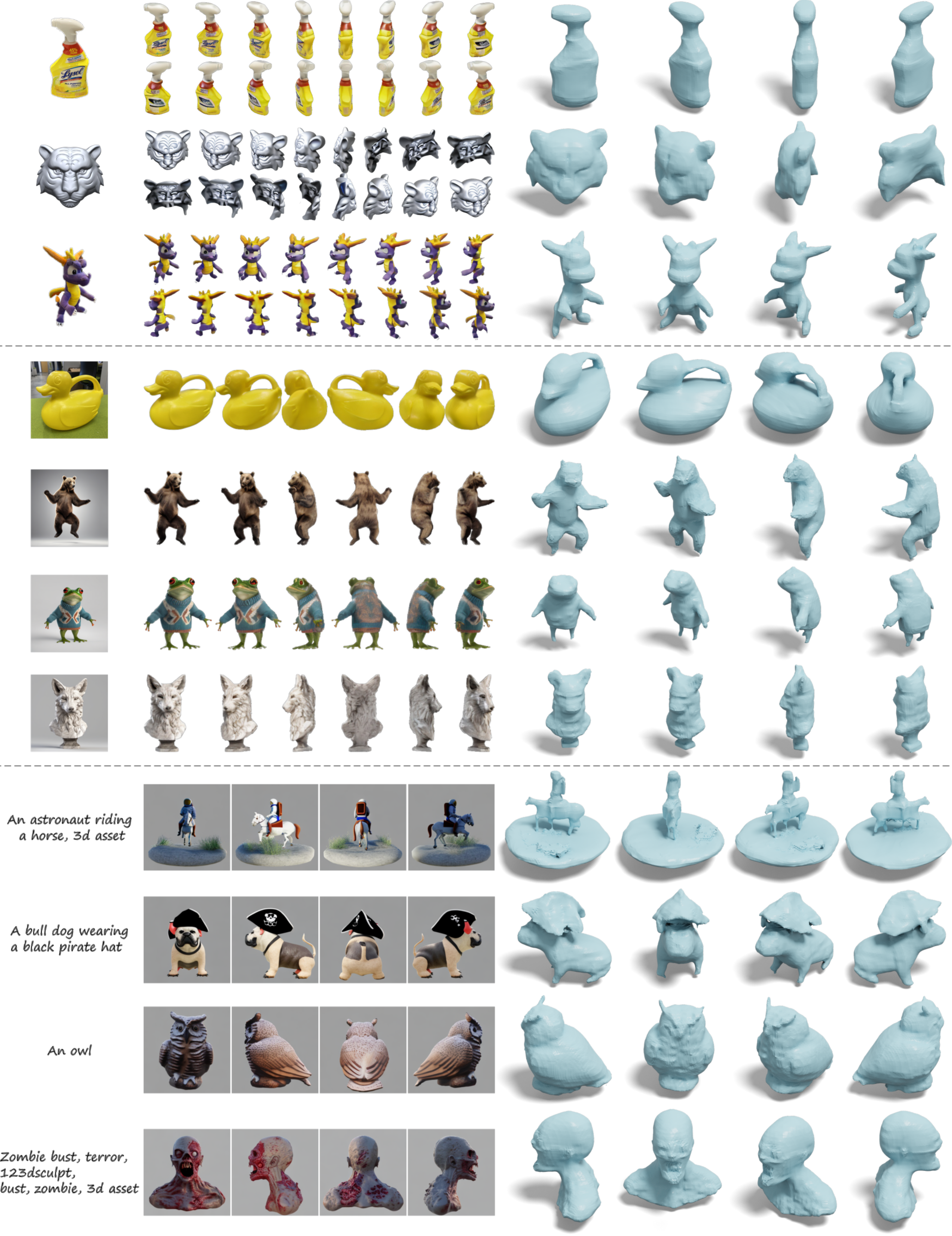}
    \caption{Generalizability test of {\mvd}. From left to right: prompt image or text, MVD images, the reconstructed shape rendered from four different angles. The MVD images of the first three examples, the next four examples, and the last four examples, are produced using SyncDreamer, Wonder3D, and MVDream respectively.
    }
    \label{fig:robust}
\end{figure*}

\paragraph{Generalizability to image-conditioned MVD models} We tested SyncDreamer and Wonder3D with some online images as conditions and used the pretrained {\mvd} to reconstruct 3D shapes from the MVD images. As illustrated in \cref{fig:robust}, our method can effectively process these inputs and produce visually pleasing shapes.

\paragraph{Generalizability to text-conditioned MVD model}
We also apply {\mvd} to the four-view images produced by the text-conditioned MVD model, MVDream~\cite{shi2023mvdream}. As \cref{fig:robust} shows, {\mvd} handles MVDream's output well and generates convincing shape geometry.

\subsection{Ablation study} \label{subsec:ablation}
To assess the effectiveness of our training scheme, we conduct an ablation study with five different settings, which differ in whether they use ground-truth images or generated images for loss supervision and whether they apply view-dependent loss or not.

\begin{enumerate}[leftmargin=*]\setlength\itemsep{1mm}
    \item[\textbf{A1}.] The network takes as input the ground-truth images of 3D data rendered at the views specified by Zero-123++. To compute the loss, we render the training 3D objects and the reconstructed mesh from 64 random views, and apply both $\mathcal{L}_{pixel}$ and $\mathcal{L}_{LPIPS}$ on all the views.
    \item[\textbf{A2}.] The setup is the same as \textbf{A1}, but we follow One-2-3-45++~\cite{liu2023oneplus} and add random and small view perturbations when fetching image features during training. We aim to test whether this perturbation can improve the robustness and better handle the inconsistency of MVD images.
    \item[\textbf{A3}.] The setup is the same as \textbf{A1}, except that the input MV images are MVD images produced by Zero-123++.
    \item[\textbf{A4}.] The setup is the same as \textbf{A3}, except that we only use the images at the reference view $v_0$ and six other views in the loss computation.
    \item[\textbf{A5}.] Our default setting, where $\mathcal{L}_{pixel}$ is applied only to view $v_0$.
\end{enumerate}

\begin{figure}[t]
    \centering
    \begin{overpic}[width=\linewidth]{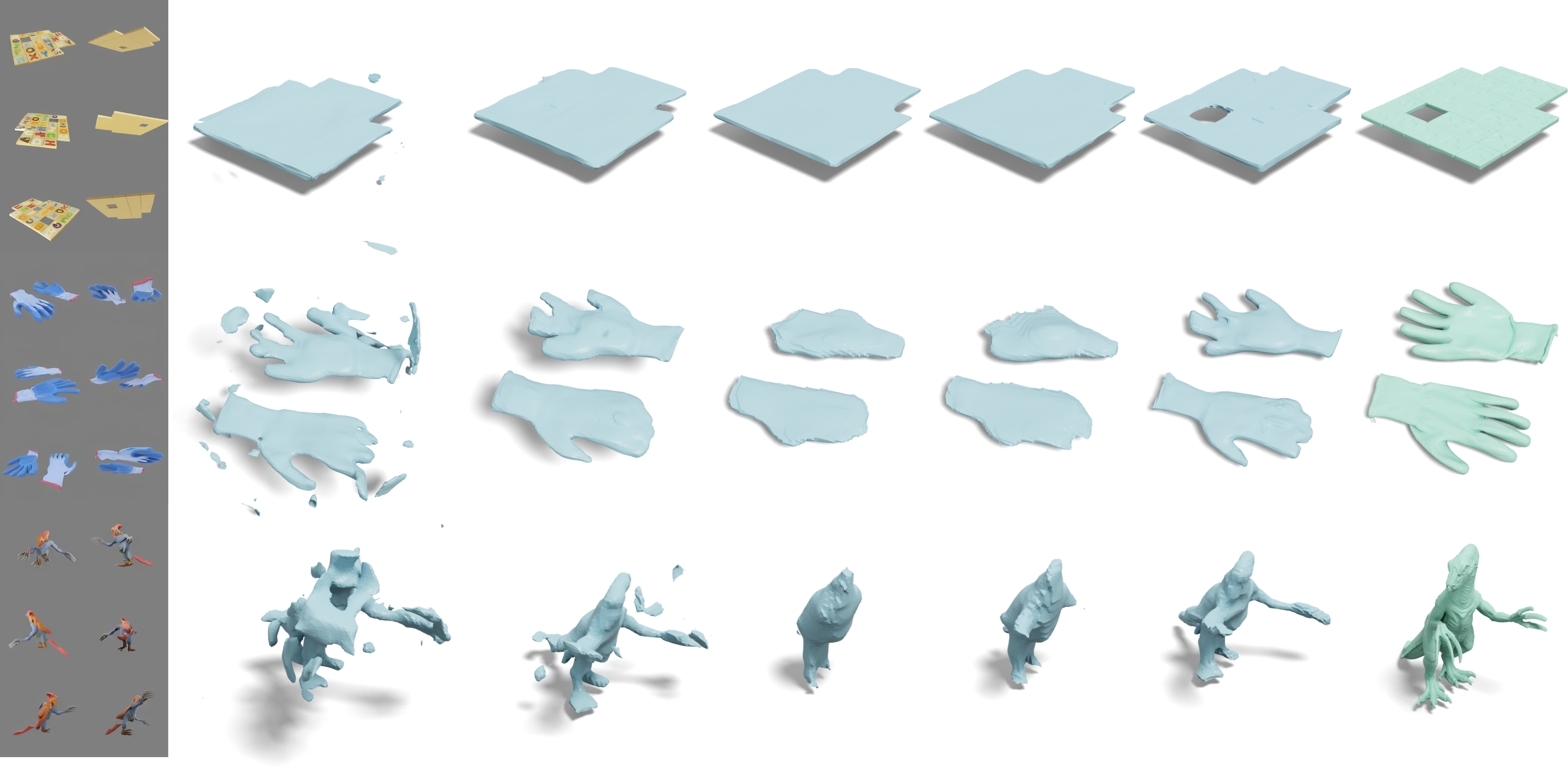}
    \put(18,-1.5){\small \textbf{A1}}
    \put(36,-1.5){\small \textbf{A2}}
    \put(50.5,-1.5){\small \textbf{A3}}
    \put(63.5,-1.5){\small \textbf{A4}}
    \put(76,-1.5){\small \textbf{A5}}
    \put(90,-1.5){\small GT}
    \end{overpic}
    \caption{Visualization of three examples reconstructed by different variants of {\mvd}. Left is the input MVD images. 
    }
    \label{fig:ablation}
\end{figure}

As shown in \cref{tab:ablation}, the geometry quality improves as the training setup approaches our default setup. Training with ground-truth render images yields the worst results, although view perturbation alleviates the problem to some degree. \textbf{A3} and \textbf{A4} show that applying $\mathcal{L}_{pixel}$ to all the views is less effective than our view-dependent training scheme.
In \cref{fig:ablation}, we also present three test cases, where our default setting \textbf{A5} clearly avoids floating geometry, reconstructs more detailed geometry such as  the empty zone of the alphabet puzzle board (upper), glove's fingers (middle), and the witch model's arm (lower).

\begin{table*}[t]
    \caption{Ablation study of the training scheme. \textbf{VP} stands for viewport perturbation. \textbf{\#v} indicates denotes the number of ground-truth views that are used in loss supervision, and \textbf{VDL} means the use of view-dependent loss. A5 is our final model.
    }\label{tab:ablation}
    \centering
    \scalebox{1}{
        \begin{tabular}{@{}cccc|*{8}{c}@{}}
            \toprule
                        & \thead{Training images} & \thead{\#v} & \thead{VDL} & \thead{CD}$\big\downarrow$ & \thead{EMD}$\big\downarrow$ & \thead{PSNR}$_d$$\big\uparrow$ & \thead{SSIM}$_d$$\big\uparrow$ & \thead{LPIPS}$_d$$\big\downarrow$ & \thead{PSNR}$_n$$\big\uparrow$ & \thead{SSIM}$_n$$\big\uparrow$ & \thead{LPIPS}$_n$$\big\downarrow$ \\ \midrule
            \textbf{A1} & GT                      & 64          & \xmark      & 4.371                      & 28.56                       & 17.22                          & 81.10                          & 25.59                             & 13.34                          & 72.04                          & 32.28                             \\
            \textbf{A2} & GT(VP)                  & 64          & \xmark      & 2.172                      & 19.95                       & 20.83                          & 86.76                          & 15.32                             & 16.29                          & 77.47                          & 21.55                             \\
            \textbf{A3} & Generated               & 64          & \xmark      & 1.256                      & 13.80                       & 23.02                          & 89.10                          & 11.74                             & 18.09                          & 79.88                          & 17.60                             \\
            \textbf{A4} & Generated               & 7           & \xmark      & 1.214                      & 14.62                       & 23.03                          & 89.09                          & 11.63                             & 18.07                          & 79.77                          & 17.47                             \\
            \textbf{A5} & Generated               & 7           & \cmark      & \textbf{1.044}             & \textbf{13.58}              & \textbf{23.31}                 & \textbf{89.20}                 & \textbf{10.80}                    & \textbf{18.28}                 & \textbf{80.01}                 & \textbf{16.34}                    \\
            \bottomrule
        \end{tabular}
    }
\end{table*}

\subsection{Limitations} \label{subsec:limitation}

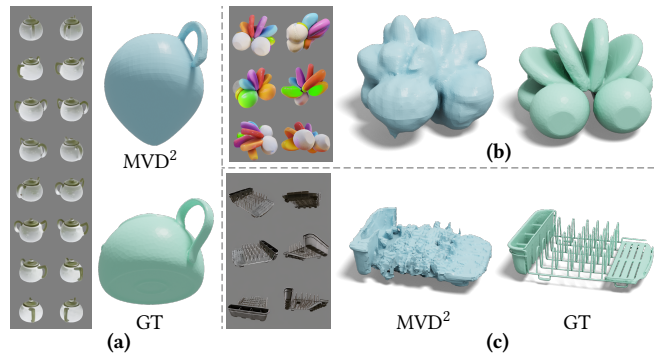
\begin{figure}[t]
    \centering
    \begin{overpic}[width=\linewidth]{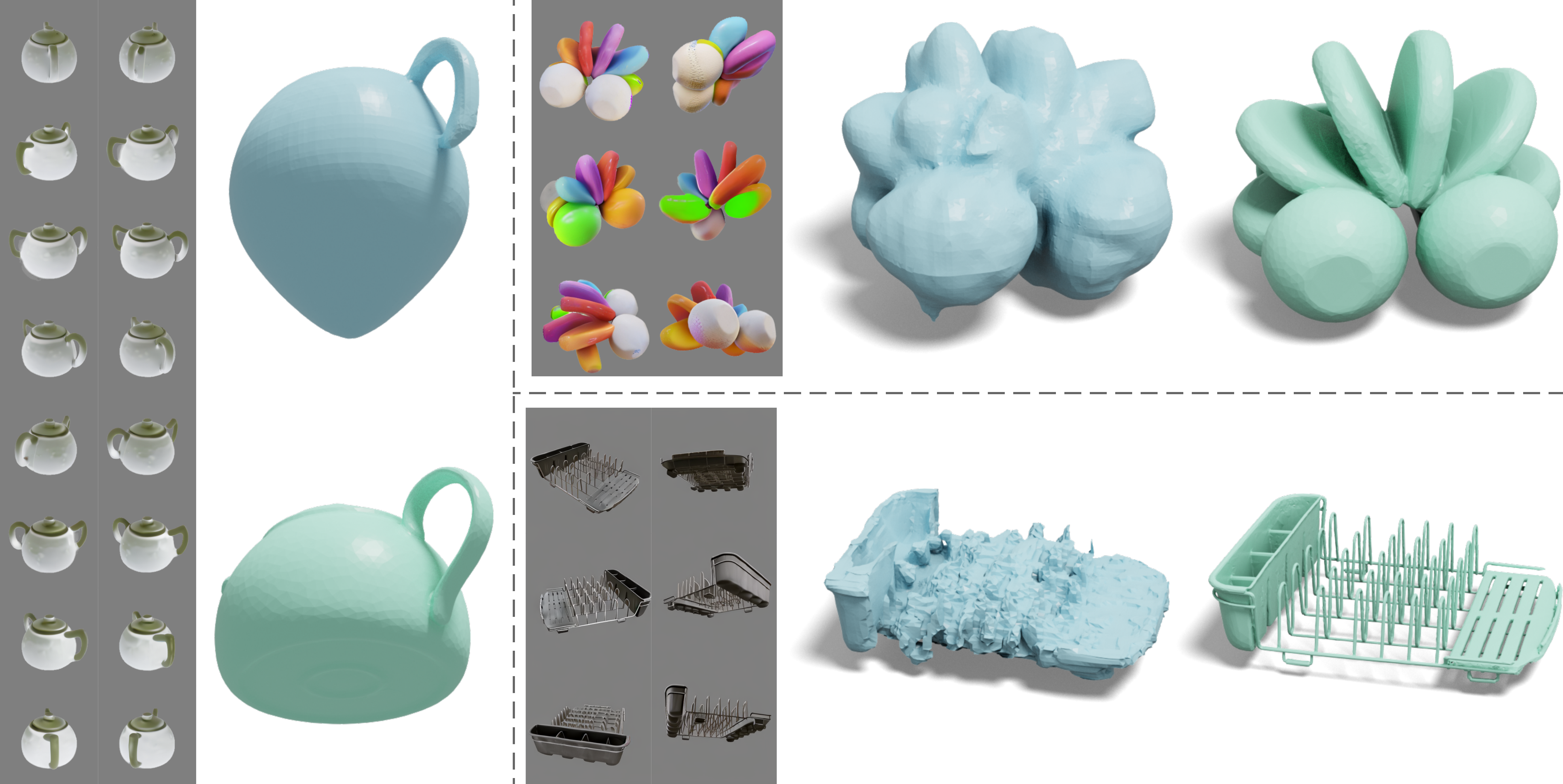}
    \put(17.5,25){\small {\mvd}}
    \put(19.5, 0){\small GT}
    \put(60,0){\small {\mvd}}
    \put(86,0){\small GT}
    \put(15,-3){\small \textbf{(a)}}
    \put(74,26.5){\small \textbf{(b)}}
    \put(74,-3){\small \textbf{(c)}}
    \end{overpic}
    \caption{Illustration of imperfect and failure reconstruction results. GTs are the 3D objects for reference. }
    \label{fig:limitation} \vspace{-3mm}
\end{figure}

Our method relies on the assumption that the input views contain sufficient information of 3D shapes. However, if a large portion of the 3D shape is invisible in all input views, the inferred 3D shape for the unseen part will deteriorate. \cref{fig:limitation}-a illustrates an example where the views generated by SyncDreamer do not include the teapot bottom. In this case, {\mvd} produces a cone-like geometry for the bottom region, which is inconsistent with human expectations.

Moreover, our method is robust to minor inconsistency of generated images, but it may fail when the generated views are highly discordant. \cref{fig:limitation}-b shows the reconstructed mesh that has a large visual discrepancy with the inconsistent input MVD images.

Additionally, our GPU memory budget limits us from using higher grid resolutions ($>80^3$) in mesh generation. Therefore, our approach cannot easily reconstruct thin geometric structures, as shown in \cref{fig:limitation}-c.

\section{Conclusion}
We propose a novel multiview 3D reconstruction method that enhances the 3D reconstruction quality from multiview depth (MVD) images. Our method leverages a carefully designed training scheme that mitigates the discrepancy between the MVD images and the 3D training data. Our {\mvd} model, which is pretrained on MVD images from Zero-123++, surpasses previous reconstruction pipelines and other 3D generation methods in terms of quality and efficiency. We conjecture that finetuning our model on the output of other MVD models will further boost their 3D reconstruction quality.

\bibliographystyle{ACM-Reference-Format}
\bibliography{src/ref}

\end{document}